\documentclass[letterpaper, 10 pt, conference]{ieeeconf}  

\IEEEoverridecommandlockouts                              

\usepackage[table]{xcolor}
\usepackage{changes}
\usepackage{authblk}
\usepackage{amsfonts}
\usepackage{amsmath,amssymb}
\usepackage{hyperref,cite}
\usepackage{cleveref}
\Crefname{figure}{Fig.}{Figs.}
\usepackage{algorithm}
\usepackage{algorithmic}
\let\labelindent\relax
\usepackage{enumitem}
\usepackage{float}
\usepackage{graphicx}
\usepackage[most]{tcolorbox}

\usepackage{booktabs}
\usepackage{threeparttable}
\usepackage{hhline}

\newtheorem{remark}{Remark}

\newcommand{\diag}{\mathop{\rm diag}\nolimits}

\newcommand{\rr}{{\mathbb R}}
\newcommand{\ba}[1]{\begin{array}{#1}}
\newcommand{\ea}{\end{array}}

\newcommand{\mr}[1]{\mathrm{#1}}

\title{\LARGE \bf
DR-MPC: Fast and Feasible Dynamics-Relaxed \\Model-Predictive Control for Legged Locomotion
}

\author{Run Wang$^{1,*}$, Alapati Tuerxun$^{1}$, Shuo Liu$^{2}$, Wei Xiao$^{3}$, J\'an Drgo\v na$^{4}$, Yilin Mo$^{1}$, Liang Wu$^{4,*}$
\thanks{$^*$Equal Contributions. Corresponding Author: Liang Wu. $^{1}$Run Wang, Alapati Tuerxun and Yilin Mo are with the Department of Automation, Tsinghua University, Beijing, China. {\tt\small \{wangrun24,alpttex25\}@mails.tsinghua.edu.cn, ylmo@tsinghua.edu.cn}. $^{2}$Shuo Liu is with the Department of Mechanical Engineering, Boston
University, Brookline, MA, USA {\tt\small liushuo@bu.edu}. $^{3}$Wei Xiao is with School of Electrical and Electronic Engineering,
Nanyang Technological University, Singapore. {\tt\small wei.xiao@ntu.edu.sg}. $^{4}$J\'an Drgo\v na and Liang Wu are with Johns Hopkins University, Baltimore, MD 21218, USA. {\tt\small \{wliang14,jdrgona1\}@jh.edu}}%
}

\begin{document}

\maketitle
\thispagestyle{empty}
\pagestyle{empty}

\begin{abstract}

    This paper presents dynamics-relaxed model predictive control (DR-MPC), a novel MPC formulation for legged locomotion, and a tailored interior-point method (IPM) solver. The formulation combines online optimization feasibility by construction with a contact-aware input parameterization. DR-MPC moves the dynamics equality and affine input constraints into quadratic penalties and retains only nonempty box constraints. The resulting box-constrained quadratic program (QP) has a block-arrow Hessian that enables the state and affine-output directions to be eliminated through a Schur complement. The solver factors only the reduced control system after swing-force elimination and contact-aligned move blocking. For the evaluated implementations using the same DR-MPC formulation, our method achieves median end-to-end MPC speedups of $16.0\times$ over HPIPM and $4.4\times$ over OSQP, with comparable locomotion performance in simulation. DR-MPC achieves a median onboard MPC end-to-end time of $4.4$ ms and is validated on a Unitree Go1 quadruped. Open-source code will be made available after publication.

\end{abstract}

\section{Introduction}
Model predictive control (MPC) has become a key tool for dynamic legged locomotion because it can continuously adapt ground reaction forces to changing commands, contacts, and disturbances \cite{di2018dynamic,bishop2025surprising}. Its effectiveness, however, depends strongly on the replanning rate. Faster replanning can improve closed-loop responsiveness, but requires the underlying optimization problem to be constructed and solved within a tight real-time budget. Onboard implementations must share computing resources with state estimation, foothold planning, and low-level control. The relevant budget therefore covers the complete MPC update, including model construction, solver-specific data preparation, and numerical optimization. At the same time, conventional hard-constrained MPC can become infeasible due to inconsistencies among state estimates, contact constraints, and prediction dynamics, particularly under large disturbances or contact transitions \cite{scokaert1999feasibility}. Legged-robot MPC therefore benefits from quadratic programs (QPs) that are \textit{fast to solve} and \textit{feasible by construction} while retaining accurate closed-loop control.

A distinctive feature of legged-robot MPC is that its prediction model is already an approximation. High-rate controllers commonly use single rigid body dynamics (SRBD) or related reduced-order models instead of full whole-body dynamics~\cite{di2018dynamic,ding2021representation,bishop2025surprising}. The SRBD model used here captures the net effect of ground reaction forces on body translation and rotation while omitting articulated-leg dynamics and contact compliance. This reduces the cost of online prediction, but leaves model discrepancies that cannot be eliminated by solving the optimization problem more accurately. However, standard MPC formulations enforce the approximate prediction model exactly over the entire horizon, either explicitly through equality constraints in sparse MPC-to-QP formulations or implicitly through recursive elimination of the state variables in condensed formulations \cite{jerez2011condensed,wu2023simple}.
\begin{tcolorbox}[
        colback=gray!50,       
        colframe=black,        
        arc=2mm,
        auto outer arc,
        opacityfill=0.1,       
        boxrule=1pt           
    ]
    This motivates a design question: can penalized deviations from an approximate prediction model yield a more favorable optimization structure? We explore this trade-off without assuming that relaxation alone improves closed-loop robustness.
\end{tcolorbox}
\subsection{Related Work}

\begin{figure}[t]
    \centering
    \includegraphics[width=\columnwidth]{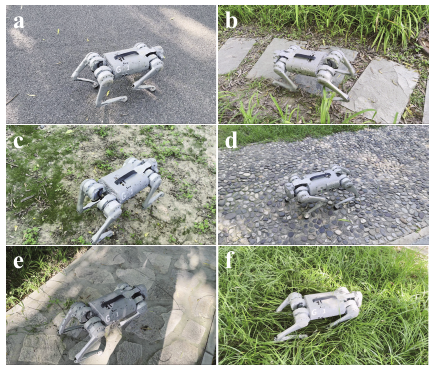}
    \vspace{-20pt}
    \caption{\textbf{DR-MPC experiment snapshots.} The quadruped walks on (a) asphalt, (b) discontinuous flagstones, (c) uneven soil with sparse vegetation, (d) cobblestones, (e) an irregular stone-paved path, and (f) dense grass.}
    \label{fig:real_world_exp_snapshots}
\end{figure}

\begin{figure*}[t]
    \centering
    \includegraphics[width=\textwidth]{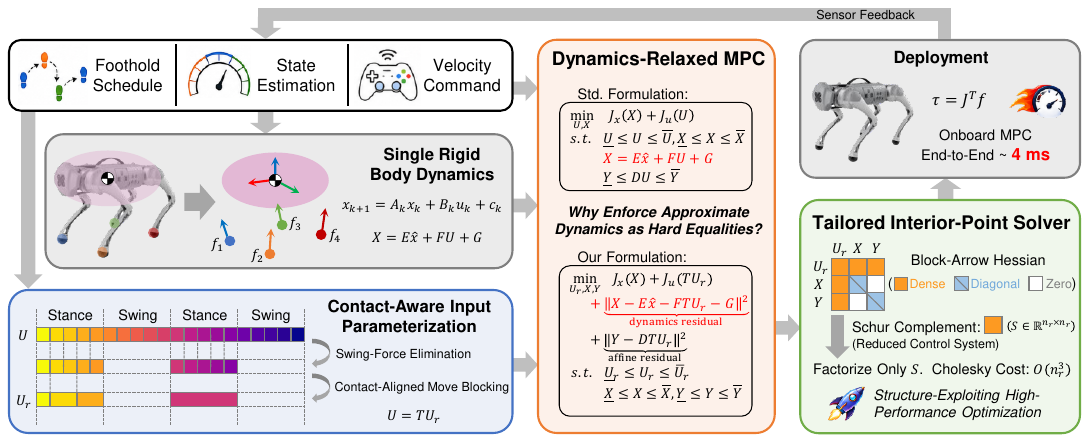}
    \vspace{-20pt}
    \caption{\textbf{Architecture of DR-MPC for legged locomotion.} The state estimate, velocity command, and planned contacts parameterize the single-rigid-body prediction model and the contact-aware input parameterization. Swing-force elimination and contact-aligned move blocking produce the reduced control variable $U_r$. Dynamics relaxation yields a box-constrained QP with a block-arrow Hessian, allowing the state and friction-output directions to be eliminated by a Schur complement. The tailored interior-point method therefore factorizes only the Schur complement $S\in\mathbb{R}^{n_r\times n_r}$, where $n_r=\dim(U_r)$, with an $O(n_r^3)$ Cholesky cost for high-frequency onboard replanning.}
    \label{fig:drmpc_architecture}
\end{figure*}

Fast MPC methods exploit sparsity, condensing, Riccati recursions, or tailored QP solvers such as HPIPM (high-performance interior-point method) and OSQP (operator splitting quadratic program)~\cite{stellato2020osqp,frison2020hpipm,verschueren2022acados,axehill2015sparsity}. Move blocking further reduces control decisions by sharing inputs across prediction stages~\cite{cagienard2007move,chen2020efficient,otta2021quadratic}.

These techniques do not ensure feasibility when dynamics and bounds conflict. Aggressive blocking can shrink the feasible set~\cite{mayne2000constrained,gondhalekar2010least,shekhar2015optimal}. Softening state constraints can restore feasibility with hard dynamics~\cite{zeilinger2014soft} and can be combined with input reduction. Our distinction is the relaxed coupling and solver structure, not feasibility or freedom from penalty tuning alone.

Koopman-BoxQP~\cite{wu2026koopman} combines dynamics relaxation and affine-constraint relaxation using auxiliary outputs with a structured box-constrained QP (BoxQP). Its tailored interior-point method (IPM) eliminates state directions through a Schur complement. DR-MPC builds on these foundations for contact-dependent legged locomotion.

\subsection{Proposed Approach}

We propose \textit{dynamics-relaxed MPC} (DR-MPC) for legged locomotion, combining BoxQP with contact-aware input parameterization and a tailored solver. DR-MPC penalizes dynamics and affine input residuals while retaining nonempty boxes, guaranteeing pointwise QP feasibility rather than exact satisfaction of the relaxed relations. The formulation still uses auxiliary outputs and penalty weights. Fig.~\ref{fig:drmpc_architecture} summarizes the onboard pipeline.

Swing-foot forces are eliminated under the nominal zero-force assumption. Stance forces use contact-aligned move blocking, with finer resolution near the horizon start. When sufficient blocks are available, every contact-mode transition receives a block boundary. The resulting BoxQP retains a block-arrow Hessian. Our predictor-corrector IPM combines direct reduced prediction assembly and local friction-block updates with Schur elimination, factorizing only the reduced control system. The factorization cost therefore depends on the reduced control dimension.

Onboard simulation benchmarks evaluate this formulation--parameterization--solver co-design. Using the same DR-MPC formulation, our implementation achieves median end-to-end speedups of $16.0\times$ over the evaluated HPIPM implementation and $4.4\times$ over OSQP, with comparable tracking. A median MPC time of $4.4$ ms supports nominal 100-Hz replanning. Hardware experiments validate locomotion over grass, slopes, flagstones, cobblestones, and other irregular terrain.

\subsection{Contributions}
The main contributions are:
\begin{itemize}
    \item We formulate DR-MPC for legged locomotion by adapting dynamics-relaxed BoxQP to contact-dependent SRBD prediction, yielding an optimization problem that is feasible by construction without explicit violation slacks.

    \item We introduce a contact-aware input parameterization that combines swing-force elimination with contact-aligned move blocking, directly reducing the effective control dimension while preserving short-horizon control authority.

    \item We develop a tailored interior-point solver that combines direct reduced assembly and local friction-block updates with Schur elimination to factor only the reduced control system.

    \item We benchmark DR-MPC against HPIPM- and OSQP-based hard-, soft-, and dynamics-relaxed MPC formulations in simulation on the onboard computer, and validate it on a physical Unitree Go1.
\end{itemize}

\section{Preliminaries}

\subsection{Single Rigid Body Dynamics}
\label{ssec:srbd}

We use the SRBD model that is standard in convex MPC for legged locomotion~\cite{di2018dynamic}. The state $x=\mathrm{col}(\Theta,p,\omega,v)\in\rr^{12}$ contains the ZYX Euler angles $\Theta=\mathrm{col}(\phi,\vartheta,\psi)$, body position $p\in\rr^3$, angular velocity $\omega\in\rr^3$, and linear velocity $v\in\rr^3$, where $\phi$, $\vartheta$, and $\psi$ denote roll, pitch, and yaw, and $\mathrm{col}$ denotes vertical stacking. The orientation is relative to the world frame, in which $p$, $\omega$, and $v$ are expressed. The input $u=\mathrm{col}(f_1,\ldots,f_4)\in\rr^{12}$ contains the world-frame ground reaction forces $f_i\in\rr^3$ of the four feet. The small-roll/pitch approximation, with coefficients frozen at the prescribed reference yaw $\psi^{\mr{ref}}$ and the gyroscopic term omitted, gives
\begin{equation}
    \label{eq:srbd}
    \begin{aligned}
        \dot\Theta & =R_z(\psi^{\mr{ref}})^\top\omega, & \dot\omega & =\hat{\mathcal I}^{-1}\sum_{i\in\mathcal C}[r_i]_\times f_i, \\
        \dot p     & =v,                               & \dot v     & =\frac{1}{m}\sum_{i\in\mathcal C}f_i+g.
    \end{aligned}
\end{equation}
Here $R_z(\psi^{\mr{ref}})\in\rr^{3\times3}$ is the rotation about the world $z$ axis by the reference yaw, $m>0$ is the body mass, $g\in\rr^3$ is the gravity vector, $r_i\in\rr^3$ is the world-frame position of foot $i$ relative to the center of mass, $[r_i]_\times$ is the skew-symmetric matrix satisfying $[r_i]_\times f_i=r_i\times f_i$, and $\mathcal C$ is the planned contact set. The body-frame inertia $\mathcal I_b\succ0$ is mapped to the world frame as $\hat{\mathcal I}=R_z(\psi^{\mr{ref}})\mathcal I_bR_z(\psi^{\mr{ref}})^\top$. Swing feet exert zero ground reaction force.

Writing~\eqref{eq:srbd} as $\dot x=A_cx+B_cu+c$, where $A_c$, $B_c$, and $c$ are the continuous-time state matrix, input matrix, and affine term, the frozen linear model satisfies $A_c^2=0$. Its zero-order-hold discretization over the prediction interval $\Delta t>0$ is
\begin{equation}
    \label{eq:zoh}
    \begin{aligned}
        A_d & =I+A_c\Delta t,                                   & B_d & =\left(I\Delta t+\frac{\Delta t^2}{2}A_c\right)B_c, \\
        c_d & =\left(I\Delta t+\frac{\Delta t^2}{2}A_c\right)c.
    \end{aligned}
\end{equation}
Here $I$ is the identity matrix of compatible dimension, and $A_d$, $B_d$, and $c_d$ are the discrete-time model matrices. Allowing these matrices to vary across prediction stages gives $x_{k+1}=A_kx_k+B_ku_k+c_k$. Stacking this model over a prediction horizon of $N$ stages gives
\begin{equation}
    \label{eq:stackedprediction}
    X=E\hat x+FU+G,
\end{equation}
where $X=\mathrm{col}(x_1,\ldots,x_N)\in\rr^{12N}$ and $U=\mathrm{col}(u_0,\ldots,u_{N-1})\in\rr^{12N}$ are the stacked state and input, $\hat x=x_0$ is the measured initial state, and $E\in\rr^{12N\times12}$, $F\in\rr^{12N\times12N}$, and $G\in\rr^{12N}$ are the stacked initial-state, input, and affine prediction operators, respectively. These operators are updated online with the reference yaw, footholds, and planned contacts, which makes the prediction model linear time-varying.

\subsection{Standard MPC Formulations}
\label{ssec:hardqp}

The standard hard-constrained formulation (Std-Hard) treats both the SRBD model and the specified state and force limits as hard constraints. The implementation used in our comparison solves
\begin{subequations}
    \label{eq:hardqp}
    \begin{align}
        \min_{X,U}\quad  & \frac12\sum_{k=0}^{N-1} \left(\|x_{k+1}-x_{k+1}^{\mr{ref}}\|_Q^2+w_u\|u_k\|^2\right) \label{eq:hardqp-cost} \\
        \text{s.t.}\quad & x_{k+1}=A_kx_k+B_ku_k+c_k \label{eq:hardqp-dyn}                                                             \\
                         & \underline y_k\le D_ku_k\le\overline y_k \label{eq:hardqp-uaff}                                             \\
                         & \underline x_{k+1}\le x_{k+1}\le\overline x_{k+1} \label{eq:hardqp-xbox}                                    \\
                         & \underline u_k\le u_k\le\overline u_k, \qquad k=0,\ldots,N-1, \label{eq:hardqp-ubox}                        \\
                         & x_0=\hat x. \label{eq:hardqp-init}
    \end{align}
\end{subequations}
Here $x_{k+1}^{\mr{ref}}$ is the reference state, $Q\succeq0$ is the diagonal state-tracking weight, $w_u>0$ is the force-regularization weight, and $\|a\|_Q^2\triangleq a^\top Qa$. The matrix $D_k\in\rr^{16\times12}$ maps the contact forces to the 16-dimensional friction-pyramid output $D_ku_k$, with four signed inequalities per foot. Underlines and overlines denote componentwise lower and upper bounds, respectively. The force bounds also prescribe a near-zero force for swing feet. This sparse stagewise form is well suited to Riccati-based optimal control problem (OCP) solvers such as HPIPM~\cite{frison2020hpipm}. It can nevertheless become infeasible when the state bounds, contact constraints, and prediction dynamics are mutually inconsistent.

Our soft-constrained baseline (Std-Soft) adds a nonnegative slack $\delta_k\in\rr^{12}$ only to the state bounds for $k=1,\ldots,N$,
\begin{equation}
    \label{eq:softstate}
    \underline x_k-\delta_k\le x_k\le\overline x_k+\delta_k, \qquad \delta_k\ge0,
\end{equation}
and penalizes $\frac12 Z_s\|\delta_k\|^2+z_s\mathbf1^\top\delta_k$, where $Z_s>0$ and $z_s>0$ are the quadratic and linear slack weights and $\mathbf1$ is the all-ones vector of compatible dimension. Dynamics, force bounds, and friction-pyramid constraints remain hard, following standard MPC practice of enforcing physical input constraints while softening state limits. Since the hard force set is nonempty, any admissible input sequence defines a state trajectory, and state slacks absorb all bound violations. The soft-constrained QP is therefore feasible for any state estimate and prescribed contact schedule. These serve as references for the DR-MPC formulation introduced next.

\section{Dynamics-Relaxed MPC}
\label{sec:drmpc}

Motivated by the reduced-order nature of SRBD, DR-MPC specializes the dynamics-relaxed BoxQP framework~\cite{wu2026koopman} to legged locomotion. First, rather than enforcing approximate dynamics as hard equalities, it penalizes dynamics and affine friction-output residuals quadratically while retaining only nonempty box constraints. This guarantees pointwise QP feasibility without slack variables. Second, the resulting structure is particularly well suited to a contact-aware input parameterization. The stacked input can be reparameterized directly through $U=TU_r$, where $T$ expands the reduced input $U_r$, while preserving the block-arrow Hessian. This parameterization decreases the dimension $n_r=\dim(U_r)$ of the Schur system factorized by our solver, whose Cholesky cost scales as $O(n_r^3)$.

Legged locomotion offers two natural dimension-reduction mechanisms. Swing-foot forces are nominally zero under the contact schedule and can be eliminated, while stance forces can be parameterized more finely near the horizon start and more coarsely farther ahead. This nonuniform allocation suits receding-horizon control, where only the initial portion of the plan is executed before replanning.

As summarized in Fig.~\ref{fig:drmpc_architecture}, the contact-aware input parameterization combines swing-force elimination with contact-aligned move blocking. The tailored solver that exploits the resulting reduced structure is developed in \Cref{sec:solver}.

\subsection{Dynamics-Relaxed BoxQP Formulation}
\label{ssec:relax}

Let $w=E\hat x+G$ denote the free response in~\eqref{eq:stackedprediction}, and let $D=\mathrm{blkdiag}(D_0,\ldots,D_{N-1})\in\rr^{16N\times12N}$ be the stacked friction-pyramid map, where $\mathrm{blkdiag}$ denotes block-diagonal assembly. We introduce $Y=\mathrm{col}(y_0,\ldots,y_{N-1})\in\rr^{16N}$ as an auxiliary copy of the stacked output $DU$. The force regularization used by DR-MPC is
\begin{equation}
    \label{eq:forcecost}
    \begin{aligned}
        J_u(U)={} & \frac{w_u}{2}\sum_{k=0}^{N-1}\|u_k\|^2 +\frac{w_{\Delta u}}{2}\|u_0-u_{\mr{prev}}\|^2 \\
                  & +\frac{w_{\Delta u}}{2}\sum_{k=1}^{N-1}\|u_k-u_{k-1}\|^2,
    \end{aligned}
\end{equation}
where $u_{\mr{prev}}$ is the first-step force from the previous MPC solution and $w_{\Delta u}\ge0$ weights successive force variations.
The complete formulation is
\begin{subequations}
    \label{eq:relaxcost}
    \begin{align}
        \min_{U,X,Y}\quad
                         & \frac12\|X-X^{\mr{ref}}\|_{W_x}^2+J_u(U)\notag                                                \\
                         & \quad+\frac{\rho_d}{2}\|X-w-FU\|^2 +\frac{\rho_f}{2}\|Y-DU\|^2 \label{eq:relaxcost-objective} \\
        \text{s.t.}\quad & \underline U\le U\le\overline U \label{eq:relaxcost-ubox}                                     \\
                         & \underline X\le X\le\overline X \label{eq:relaxcost-xbox}                                     \\
                         & \underline Y\le Y\le\overline Y. \label{eq:relaxcost-box}
    \end{align}
\end{subequations}
Here $X^{\mr{ref}}=\mathrm{col}(x_1^{\mr{ref}},\ldots,x_N^{\mr{ref}})$ is the stacked reference, $W_x=\mathrm{blkdiag}(Q,\ldots,Q)\succeq0$ is its tracking weight, and $\rho_d>0$ and $\rho_f>0$ penalize the dynamics and friction residuals. The stacked bounds collect their stagewise counterparts. The state bounds are centered on the reference trajectory. The input bounds impose normal-force and tangential-force limits for stance feet, while the bounds on $Y$ encode the signed friction pyramid. The contact-aware input parameterization is discussed in \Cref{ssec:moveblock}.

Let $R_U\succ0$ denote the Hessian of~\eqref{eq:forcecost}. With the decision vector $\chi=\mathrm{col}(U,X,Y)$ and the auxiliary block $\xi=\mathrm{col}(X,Y)$, the Hessian of~\eqref{eq:relaxcost} has the block-arrow form
\begin{equation}
    \label{eq:blockarrow}
    P=\begin{bmatrix}
        P_{UU}        & P_{U\xi}    \\
        P_{U\xi}^\top & \Lambda_\xi
    \end{bmatrix},
\end{equation}
where
\begin{equation}
    \begin{aligned}
        P_{UU}      & =R_U+\rho_dF^\top F+\rho_fD^\top D,                       \\
        P_{U\xi}    & =\begin{bmatrix}-\rho_dF^\top&-\rho_fD^\top\end{bmatrix}, \\
        \Lambda_\xi & =\mathrm{blkdiag}(W_x+\rho_d I,\rho_f I).
    \end{aligned}
\end{equation}
The matrix $R_U$ is block tridiagonal because the force-variation penalty couples consecutive inputs. The trailing block $\Lambda_\xi$ is diagonal because $W_x$ is diagonal and the $X$ and $Y$ variables do not couple with one another.

\begin{remark}[Feasibility]
    \label{rem:properties}
    If every interval in~\eqref{eq:relaxcost-ubox}--\eqref{eq:relaxcost-box} is nonempty, their Cartesian product is nonempty for every state estimate and planned contact schedule. DR-MPC is therefore feasible as an optimization problem without slack variables. This statement does not claim exact satisfaction of the relaxed dynamics or friction relations. Their residuals are controlled by $\rho_d$ and $\rho_f$. For fixed prediction matrices, contact schedule, control map, bounds, and weights, the strongly convex BoxQP has a Lipschitz solution map with respect to its linear coefficient~\cite{wu2026koopman}.
\end{remark}

\subsection{Contact-Aware Input Parameterization}
\label{ssec:moveblock}

The contact-aware input parameterization shown in \Cref{fig:drmpc_architecture} combines swing-force elimination and contact-aligned move blocking in one sparse control map
\begin{equation}
    \label{eq:controlmap}
    U=TU_r.
\end{equation}
Here $U_r\in\rr^{n_r}$ is the reduced control vector, $n_r$ is its dimension, and $T\in\rr^{12N\times n_r}$ expands it into the full input $U$. The subscript $r$ identifies reduced control coordinates, while overlines and underlines denote bounds. Swing-force elimination omits inactive swing slots under the nominal zero-force assumption. Contact-aligned move blocking assigns each remaining column of $T$ to one force component of one stance foot and copies it across a control block. Let $N_u$ denote the selected number of blocks and $N_c$ the number of maximal constant-contact segments in the horizon. When $N_u\ge N_c$, every contact-mode switch receives a block boundary, preventing blocks from spanning touchdown or liftoff. Additional boundaries refine the horizon start.

For contact-aligned blocks, shared input bounds are the intersection of identical, nonempty force intervals, so the implementation uses the first slot's bounds. Under-resolved $N_u=3,4$ ablations retain the block-start contact mode and bounds across later switches and need not preserve stagewise contact consistency.

Substituting~\eqref{eq:controlmap} into~\eqref{eq:blockarrow} gives
\begin{equation}
    \label{eq:reducedP}
    \begin{aligned}
        P_{rr}   & =T^\top R_UT+\rho_dF_r^\top F_r +\rho_fD_r^\top D_r, \\
        P_{r\xi} & =T^\top P_{U\xi},                                    \\
        F_r      & =FT,                                                 \\
        D_r      & =DT.
    \end{aligned}
\end{equation}
The reduced prediction matrix $F_r$ is constructed directly inside the prediction recursion without forming the full matrix $F$. The sparse friction map $D_r$ is assembled directly at the same reduced width. Setting $T=I$ recovers the unreduced formulation. Eliminating fixed zero-force variables preserves the nominal stance-force decision space, whereas move blocking restricts stance-force profiles. Fig.~\ref{fig:tradeoff_between_computation_and_performance} evaluates the resulting trade-off, including under-resolved $N_u=3,4$ settings that are not always contact-aligned.

A fixed dimension is maintained for each gait, sized by the maximum number of independent control variables over one gait cycle. Inactive columns are padded with swing slots bounded by $[-1,1]$ N and excluded from the SRBD prediction. These numerical slots can still affect force regularization, so their removal is not an exact equivalence to the numerically relaxed full-input QP. The workspace is resized only when the gait changes.

\section{Tailored Interior-Point Solver}
\label{sec:solver}

We adapt the BoxQP IPM of~\cite{wu2026koopman} to \Cref{sec:drmpc}, combining its Schur elimination with direct reduced assembly and local friction updates that exploit the legged contact structure.

Let $q=\mathrm{col}(U_r,X,Y)$ collect the reduced physical variables, with componentwise bounds $\underline q$ and $\overline q$. Define the bound center $q_c=(\overline q+\underline q)/2$ and the positive diagonal half-range matrix $S_q=\diag((\overline q-\underline q)/2)$. Here $\diag$ constructs a diagonal matrix from a vector or extracts the diagonal of a matrix. Assuming strictly positive interval widths, the affine change of variables $q=q_c+S_q\zeta$ maps the physical bounds onto $[-\mathbf1,\mathbf1]$. The reduced DR-MPC problem becomes
\begin{equation} \label{eq:boxqp}
    \begin{aligned}
        \min_{\zeta} & ~\frac12\zeta^\top H\zeta+h^\top\zeta, \\
        \text{s.t.}  & ~-\mathbf1\le\zeta\le\mathbf1,
    \end{aligned}
\end{equation}
where $\zeta\in\rr^n$ is the normalized decision vector, $n=\dim(\zeta)$, $H\succ0$ and $h$ are the transformed Hessian and linear coefficient, and $\mathbf1\in\rr^n$ is the all-ones vector. The partition $\zeta=\mathrm{col}(\zeta_r,\zeta_\xi)$, with $\zeta_\xi=\mathrm{col}(\zeta_X,\zeta_Y)$, separates the reduced-control, state, and friction-output coordinates. The Hessian retains the block-arrow structure in~\eqref{eq:blockarrow}.

We solve~\eqref{eq:boxqp} with a feasible Mehrotra predictor-corrector interior-point method~\cite{mehrotra1992implementation}. Let $\lambda^+>0$ and $\lambda^->0$ be the upper-bound and lower-bound multipliers, with corresponding slacks $s^+=\mathbf1-\zeta>0$ and $s^-=\mathbf1+\zeta>0$. Initialization enforces stationarity, $H\zeta+h+\lambda^+-\lambda^-=0$, which the Newton updates preserve in exact arithmetic. Eliminating the dual and slack directions gives
\begin{equation}
    \label{eq:reduced}
    (H+\Sigma)\Delta\zeta=b,
\end{equation}
where $\Delta\zeta$ is the primal search direction, $\Sigma=\diag(\lambda^+\oslash s^+ +\lambda^-\oslash s^-)$, and $\oslash$ denotes elementwise division. For complementarity right-hand sides $r_+,r_-$, stationarity gives $b=-r_+\oslash s^+ +r_-\oslash s^-$. The diagonal matrix $\Sigma\succ0$ contains the barrier terms. The predictor and corrector steps share the same coefficient matrix, so its factorization is reused within each IPM iteration. Although BoxQPs admit execution-time-certified algorithms~\cite{wu2025direct,11313851,wu2026n3QP}, the Mehrotra implementation evaluated here has no certified execution-time bound.

\subsection{Reduced Newton System}
Partition~\eqref{eq:reduced} conformably with $\zeta=\mathrm{col}(\zeta_r,\zeta_\xi)$. Because $H$ inherits the block-arrow structure of~\eqref{eq:blockarrow}, the system takes the form
\begin{equation}
    \label{eq:part}
    \begin{bmatrix}
        H_{rr}+\Sigma_r & H_{r\xi}              \\
        H_{r\xi}^\top   & H_{\xi\xi}+\Sigma_\xi
    \end{bmatrix}
    \begin{bmatrix} \Delta\zeta_r \\ \Delta\zeta_\xi \end{bmatrix}
    = \begin{bmatrix} b_r \\ b_\xi \end{bmatrix}.
\end{equation}
Here $H_{rr}$, $H_{r\xi}$, and $H_{\xi\xi}$ are the reduced-control, coupling, and auxiliary diagonal blocks of $H$. The quantities $\Sigma_r$, $\Sigma_\xi$, $b_r$, and $b_\xi$ are the conformal partitions of $\Sigma$ and $b$. The matrix $H_{\xi\xi}+\Sigma_\xi$ is diagonal because both its terms are diagonal. The normalized auxiliary variables $\zeta_\xi$ couple to $\zeta_r$ but not to one another.

This structure admits a Schur complement onto the control block alone. Eliminating $\Delta\zeta_\xi$ yields
\begin{equation}
    \label{eq:schur}
    \begin{aligned}
        S\,\Delta\zeta_r & = b_r - H_{r\xi}\,(H_{\xi\xi}+\Sigma_\xi)^{-1}\,b_\xi,                               \\[2pt]
        S                & \triangleq H_{rr} + \Sigma_r - H_{r\xi}\,(H_{\xi\xi}+\Sigma_\xi)^{-1} H_{r\xi}^\top.
    \end{aligned}
\end{equation}
The Schur complement $S\in\rr^{n_r\times n_r}$ is symmetric positive definite. As highlighted in \Cref{fig:drmpc_architecture}, only $S$ requires factorization, by Cholesky at cost $O(n_r^3)$. The eliminated variables are then recovered by diagonal back-substitution,
\begin{equation}
    \label{eq:backsub}
    \Delta\zeta_\xi = (H_{\xi\xi}+\Sigma_\xi)^{-1}\bigl(b_\xi - H_{r\xi}^\top\,\Delta\zeta_r\bigr).
\end{equation}
The diagonal inverse uses elementwise reciprocals. Input reduction lowers the $O(n_r^3)$ factorization cost, while the dense state contribution to Schur assembly costs $O(n_Xn_r^2)$, where $n_X=12N$. Total update time also depends on QP construction, back-substitution, and iteration count.

\subsection{Initialization and Warm Start}
The cold start sets $\zeta^0=0$ and uses
\begin{equation}
    \label{eq:coldstart}
    \begin{aligned}
        \eta            & =\max(\|h\|_\infty,1),\qquad s^{\pm,0}=\mathbf1, \\
        \lambda^{\pm,0} & =\eta\mathbf1\mp\frac12h.
    \end{aligned}
\end{equation}
Here $\eta$ scales the initial multipliers. This point is strictly feasible. The solver terminates when the duality measure
$\mu=((\lambda^+)^\top s^+ +(\lambda^-)^\top s^-)/(2n)$ satisfies $\mu\le\epsilon$, where $\epsilon>0$ is the convergence tolerance.

For warm starting, the previous solution is stored in the full physical coordinates $(U,X,Y)$. It is shifted by the number of MPC stages elapsed since the preceding solve, and the terminal entries are repeated. The shifted full input is then projected onto the current contact-aware input map by averaging the force slots represented by each column of $T$. Finally, the complete guess is scaled and clipped to the strict interior of $[-\mathbf1,\mathbf1]$. For this $\zeta^0$, we initialize $\eta$ and $\lambda^{\pm,0}$ as in~\eqref{eq:coldstart} with $h$ replaced by $H\zeta^0+h$, and set $s^{\pm,0}=\mathbf1\mp\zeta^0$. This preserves stationarity and strict feasibility when the blocking pattern changes.

Algorithm~\ref{alg:tailored_ipm} summarizes the solve. Write $z_{\mr{IPM}}=(\zeta,\lambda^+,\lambda^-,s^+,s^-)$ and let $\Delta z_{\mr{IPM}}$ collect their search directions. The operator $\alpha_\tau$ gives a step capped at one that keeps slacks and multipliers positive, with fraction-to-the-boundary parameter $\tau\in(0,1)$. Superscripts $\mr a$ and $\mr c$ denote affine-scaling and corrector quantities. The affine step $\alpha_{\mr a}$ produces the trial duality measure $\mu_{\mr a}$. The centering parameter is $\sigma\in[0,1]$, $\odot$ denotes elementwise multiplication, and $K_{\max}$ is the iteration limit. The operator $\operatorname{dir}(r_+,r_-,L)$ obtains $\Delta\zeta$ through~\eqref{eq:schur}--\eqref{eq:backsub} using the Cholesky factor $L$, then recovers $\Delta s^\pm=\mp\Delta\zeta$ and $\Delta\lambda^\pm=(r_\pm\pm\lambda^\pm\odot\Delta\zeta)\oslash s^\pm$.

\begin{algorithm}[t]
    \caption{Tailored predictor-corrector IPM for DR-MPC.}
    \label{alg:tailored_ipm}
    \small
    \begin{algorithmic}[1]
        \REQUIRE Structured BoxQP data $(H_{rr},H_{r\xi},H_{\xi\xi},h)$
        \REQUIRE Optional warm start and solver parameters $(\epsilon,\tau,K_{\max})$
        \STATE Initialize $(\zeta,\lambda^+,\lambda^-,s^+,s^-)$ with~\eqref{eq:coldstart} or the shifted warm start
        \FOR{$k=0,\ldots,K_{\max}-1$}
        \STATE $\mu\gets((\lambda^+)^\top s^+ +(\lambda^-)^\top s^-)/(2n)$
        \IF{$\mu\le\epsilon$}
        \RETURN $\zeta$ with converged status
        \ENDIF
        \STATE $\Sigma\gets\diag(\lambda^+\oslash s^+ +\lambda^-\oslash s^-)$
        \STATE Assemble $S$ in~\eqref{eq:schur} with the structure-aware updates
        \STATE Compute the Cholesky factorization $S=LL^\top$
        \STATE $\Delta z_{\mr{IPM}}^{\mr a}\gets\operatorname{dir}(-\lambda^+\odot s^+,-\lambda^-\odot s^-,L)$
        \STATE $\alpha_{\mr a}\gets\alpha_\tau(\Delta z_{\mr{IPM}}^{\mr a})$ and evaluate $\mu_{\mr a}$
        \STATE $\sigma\gets\min(1,\max(0,\mu_{\mr a}/\mu))^3$
        \STATE $r_+^{\mr c}\gets-\lambda^+\odot s^+-\Delta\lambda^{+,\mr a}\odot\Delta s^{+,\mr a}+\sigma\mu\mathbf1$
        \STATE $r_-^{\mr c}\gets-\lambda^-\odot s^--\Delta\lambda^{-,\mr a}\odot\Delta s^{-,\mr a}+\sigma\mu\mathbf1$
        \STATE $\Delta z_{\mr{IPM}}\gets\operatorname{dir}(r_+^{\mr c},r_-^{\mr c},L)$
        \STATE $\alpha\gets\alpha_\tau(\Delta z_{\mr{IPM}})$
        \STATE $z_{\mr{IPM}}\gets z_{\mr{IPM}}+\alpha\Delta z_{\mr{IPM}}$
        \ENDFOR
        \RETURN $\zeta$ with iteration-limit status
    \end{algorithmic}
\end{algorithm}

\subsection{Structure-Aware Implementation}
The IPM uses a preallocated workspace and performs no heap allocation during its iterations. The dense state contribution to the Schur complement is assembled with a symmetric rank-$k$ update that touches only the lower triangle. Let $H_{rY}$ denote the coupling between reduced controls and friction outputs in the normalized Hessian, and let $H_{YY}$ and $\Sigma_Y$ be the friction-output diagonal blocks of $H_{\xi\xi}$ and $\Sigma_\xi$. With $M_Y=H_{YY}+\Sigma_Y$, the friction contribution is a sum of local stage-foot updates,
\begin{equation}
    \label{eq:frictionblocks}
    H_{rY}M_Y^{-1}H_{rY}^\top
    =\sum_{k=0}^{N-1}\sum_{i=1}^{4}
    C_{k,i}M_{Y,k,i}^{-1}C_{k,i}^\top,
\end{equation}
where $M_{Y,k,i}\in\rr^{4\times4}$ is the diagonal block of $M_Y$ for foot $i$ at stage $k$, and $C_{k,i}\in\rr^{n_r\times4}$ contains the corresponding columns of $H_{rY}$. Each $C_{k,i}$ has at most three nonzero control rows. Equation~\eqref{eq:frictionblocks} replaces one dense product by independent small updates. The four weighted output columns yield a local control update of size at most $3\times3$. Terms sharing a blocked force accumulate in a contiguous buffer before subtraction from $S$. Only lower-triangular entries are accumulated, exploiting the update's symmetry.

Recurring contact patterns reuse cached control maps and stage-foot index lists, while numerical prediction matrices and barrier weights are updated separately.

The direct reduced prediction construction in \Cref{ssec:moveblock} avoids forming the unreduced $F$ and $D$ matrices. The implementation uses BLASFEO (Basic Linear Algebra Subroutines for Embedded Optimization) for the resulting dense operations. Together, these choices concentrate the factorization and most dense work in the reduced control dimension selected by the contact-aware input parameterization.

\section{Results}
\label{sec:results}

We evaluate the contact-aware input parameterization, compare MPC formulations and solvers in simulation, and validate the complete DR-MPC system on a physical robot.

\subsection{Experimental Setup}
Experiments use a Unitree Go1 equipped with an NVIDIA Jetson Xavier NX. State estimation, MPC construction and solution, foothold planning, and low-level control all run onboard. Comparisons vary the formulation, input parameterization, and solver while keeping the model, estimator, planner, and task fixed.

All formulations share the SRBD model, reference generator, and tracking and force-magnitude weights. The reference integrates commanded planar velocity and yaw rate at a body height of $0.26$ m. Std-Hard retains all constraints. Std-Soft relaxes only state bounds. DR-MPC penalizes dynamics and friction residuals and adds force-variation regularization. Table~\ref{tab:go1-drmpc-parameters} lists the parameters. Standard baselines use full inputs. Unless otherwise stated, DR-MPC uses swing-force elimination, five contact-aligned blocks, and $\epsilon=10^{-3}$. Cross-formulation timings compare complete configurations, not dynamics relaxation alone.

\begin{table}[t]
    \caption{Parameters of DR-MPC and the standard hard-constrained and soft-constrained MPC baselines for the Unitree Go1 experiments.}
    \label{tab:go1-drmpc-parameters}
    \centering
    \begin{threeparttable}
        \setlength{\tabcolsep}{3pt}
        \begin{tabular}{p{0.32\columnwidth}p{\dimexpr0.68\columnwidth-4\tabcolsep\relax}}
            \toprule
            \rowcolor{black!10}
            \textbf{Shared Horizon}                    & \textbf{Value}                           \\
            Prediction horizon $N$                     & $20$                                     \\
            Prediction interval $\Delta t$             & $0.02$ s                                 \\
            MPC replanning period                      & $0.01$ s ($100$ Hz)                      \\
            \midrule
            \rowcolor{black!10}
            \textbf{Shared Cost Weights}               & \textbf{Value}                           \\
            State tracking $\diag(Q)$\tnote{$\ast$}    & $[250,120,60,10,10,800,10,4,8,20,20,30]$ \\
            Force magnitude $w_u$                      & $1\times10^{-5}$                         \\
            \midrule
            \rowcolor{black!10}
            \textbf{DR-MPC Weights}                    & \textbf{Value}                           \\
            Force variation $w_{\Delta u}$             & $5\times10^{-5}$                         \\
            Dynamics relaxation $\rho_d$               & $1\times10^{4}$                          \\
            Friction relaxation $\rho_f$               & $1\times10^{2}$                          \\
            \midrule
            \rowcolor{black!10}
            \textbf{Std-Hard Weights}                  & \textbf{Value}                           \\
            Additional penalties                       & None                                     \\
            \midrule
            \rowcolor{black!10}
            \textbf{Std-Soft Weights}                  & \textbf{Value}                           \\
            State box slack $(Z_s,z_s)$                & $(1\times10^{4},1\times10^{3})$          \\
            \midrule
            \rowcolor{black!10}
            \textbf{Shared State Bounds}               & \textbf{Value}                           \\
            Half widths $x_{\text{dev}}$\tnote{$\ast$} & $[0.15,0.3,1,1,1,0.15,4,4,4,3,3,2]$      \\
            \midrule
            \rowcolor{black!10}
            \textbf{Shared Force Bounds}               & \textbf{Value}                           \\
            Friction coefficient $\mu_f$               & $0.4$                                    \\
            Stance tangential force                    & $[-40,40]$ N                             \\
            Stance normal force                        & $[2,100]$ N                              \\
            Swing force                                & $[-1,1]$ N                               \\
            \bottomrule
        \end{tabular}
        \begin{tablenotes}[flushleft]
            \footnotesize
            \item[$\ast$] The entries follow the state ordering $x=\mr{col}(\Theta,p,\omega,v)$, where $\Theta=(\phi,\vartheta,\psi)$ and the components of $p$, $\omega$, and $v$ are ordered along the $x$, $y$, and $z$ axes.
            \item The shared entries apply to DR-MPC, Std-Hard, and Std-Soft.
        \end{tablenotes}
    \end{threeparttable}
\end{table}

\subsection{Effects of Contact-Aware Input Parameterization}

\begin{figure}[t]
    \centering
    \includegraphics[width=\columnwidth]{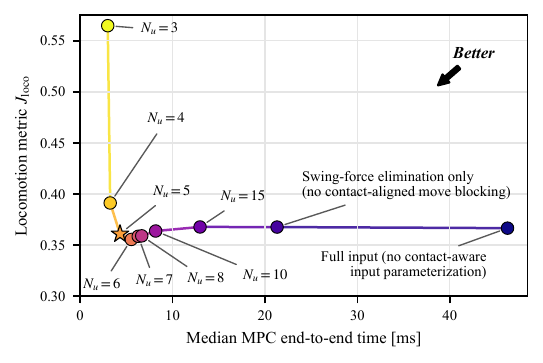}
    \vspace{-20pt}
    \caption{\textbf{Effect of the contact-aware input parameterization.} Each point gives the median locomotion metric and MPC end-to-end time under the same command sequence. Labels $N_u$ indicate swing-force elimination and $N_u$ move blocks. The $N_u=3,4$ settings can violate contact alignment. The star marks the selected configuration, and lower-left points are preferable.}\label{fig:tradeoff_between_computation_and_performance}
\end{figure}

We evaluate input parameterization in MuJoCo on the onboard processor. All simulations use fixed time steps and $10$ ms simulation-time MPC updates, without injecting measured computation delays. Tracking results therefore assess control quality at a common update interval, separately from computational demand.

At each control sample $t$, we define the tracking error as $e_t=\mr{col}(e_{v_x,t},e_{v_y,t},e_{\omega_z,t},e_{h,t},e_{\phi,t},e_{\vartheta,t})$, whose entries are the longitudinal-velocity, lateral-velocity, yaw-rate, body-height, roll, and pitch errors. The $v_x$ and $v_y$ errors use the yaw-aligned frame, unlike the world-frame velocity in the prediction state. The scalar locomotion metric $J_{\text{loco},t}$ is computed as
\begin{equation}
    \begin{aligned}
        J_{\text{loco},t} & =\sqrt{\frac{1}{6}\sum_{i=1}^{6}\left(\frac{e_{t,i}}{c_{e,i}}\right)^2}, \\
        c_e               & =\mr{col}(0.20,0.20,0.30,0.05,0.10,0.10),
    \end{aligned}
\end{equation}
where $c_e$ normalizes the six errors and balances their contributions. Its components have units of m/s, m/s, rad/s, m, rad, and rad, respectively, making $J_{\text{loco},t}$ dimensionless. We report the median of $J_{\text{loco},t}$ and the median MPC end-to-end time over each experiment.

In Fig.~\ref{fig:tradeoff_between_computation_and_performance}, the full-input formulation requires $46.25$ ms with a locomotion metric of $0.367$. Swing-force elimination reaches $21.31$ ms and $0.368$, a $2.2\times$ speedup. The selected $N_u=5$ configuration reaches $4.32$ ms and $0.361$, a $10.7\times$ speedup with comparable performance. At $N_u=3$, time falls to $2.99$ ms but the metric rises to $0.565$. Settings $N_u=3,4$ reduce control resolution and can lose contact alignment, so the degradation cannot be attributed to resolution alone. These results favor $N_u=5$ for this task.
\subsection{Simulation Comparison of MPC Methods}
\begin{figure*}[t]
    \centering
    \includegraphics[width=\textwidth]{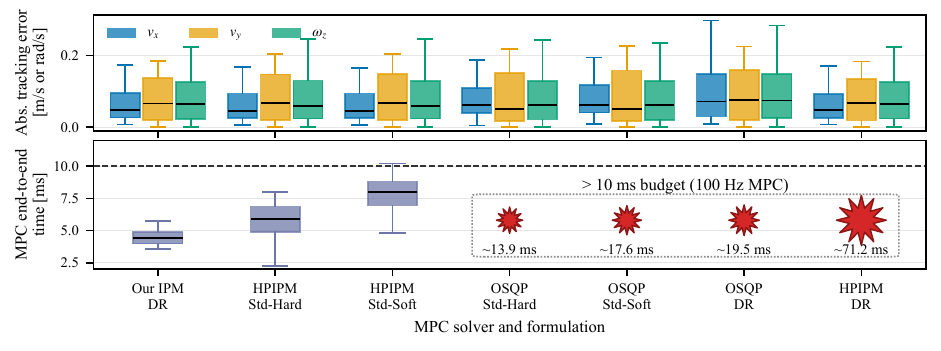}
    \vspace{-20pt}
    \caption{\textbf{Locomotion performance and MPC computation time in simulation.} Seven solver and formulation combinations follow the same command sequence. Boxes show the interquartile range, and whiskers show the $5$th and $95$th percentiles. Burst markers denote median times above the $10$ ms budget, with size indicating the overrun.}
    \label{fig:simulation_method_comparison}
\end{figure*}

Fig.~\ref{fig:simulation_method_comparison} compares seven configurations, with DR denoting DR-MPC. Tolerances balance computation and control performance empirically. All use $\epsilon=10^{-3}$ except OSQP-DR, which uses $10^{-6}$ because looser tolerances caused unstable locomotion. Different stopping criteria mean equal $\epsilon$ values need not imply equal accuracy. HPIPM uses dense QP for DR and Riccati-based OCP QP for standard formulations. Our OSQP interface rebuilds its workspace per update with primal warm starts when available. End-to-end timings include QP construction, data preparation, setup, solution, and result assembly, assessing the implemented pipelines rather than numerical iterations alone.

The tracking-error distributions largely overlap. Our method yields median absolute errors of $0.048$ m/s, $0.066$ m/s, and $0.065$ rad/s for $v_x$, $v_y$, and $\omega_z$, respectively.

Our method has the lowest median MPC end-to-end time among the evaluated configurations at $4.44$ ms. With the same DR-MPC formulation and input parameterization, it is $16.0\times$ faster than HPIPM-DR and $4.4\times$ faster than OSQP-DR. Speedups over HPIPM and OSQP are $1.8\times$ and $4.0\times$ for Std-Soft, and $1.3\times$ and $3.1\times$ for Std-Hard. Only our method and the HPIPM standard baselines have medians below $10$ ms. OSQP-Std-Hard, OSQP-Std-Soft, OSQP-DR, and HPIPM-DR require $13.9$, $17.6$, $19.5$, and $71.2$ ms. The timing advantage accompanies comparable tracking at the fixed simulation-time update interval.

\subsection{DR-MPC for Real-World Robot Locomotion}

\begin{figure}[t]
    \centering
    \includegraphics[width=\columnwidth]{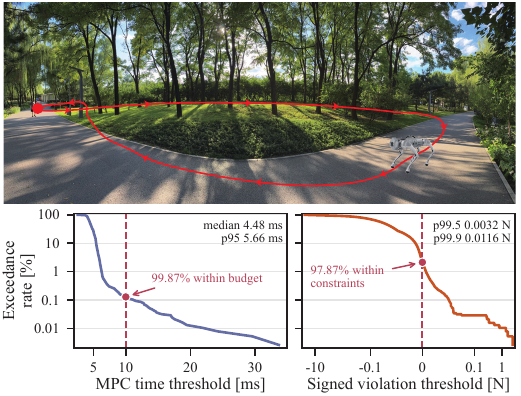}
    \vspace{-20pt}
    \caption{\textbf{Real-world locomotion and onboard MPC performance.} The panorama shows the route through grass and sloped terrain. The lower panels show empirical exceedance rates for MPC end-to-end time and the signed constraint residual of first-step optimized contact forces. Dashed lines mark $10$ ms and zero residual. The right horizontal axis uses a symmetric logarithmic scale, linear within $\pm0.01$ N.}
    \label{fig:real_world_result}
\end{figure}

The selected DR-MPC configuration completes a continuous $6.37$-min outdoor experiment on the Unitree Go1. The route in Fig.~\ref{fig:real_world_result} traverses dense grass and sloped terrain with changes in ground height, compliance, inclination, and traction.

The lower panels summarize $38{,}193$ MPC updates as percentages exceeding each threshold. Median and $95$th-percentile end-to-end times are $4.48$ ms and $5.66$ ms. Overall, $99.87\%$ meet the $10$ ms budget, with $49$ overruns and a maximum of $33.71$ ms. During asynchronous MPC computation, low-level control interpolates the latest available force trajectory. Thus, $100$ Hz denotes nominal replanning, not a strict deadline guarantee.

The signed force metric is the maximum of $|f_x|-\mu_f f_z$, $|f_y|-\mu_f f_z$, $2\,\mathrm{N}-f_z$, and $f_z-100\,\mathrm{N}$ over first-step optimized forces at planned active contacts. It checks friction-pyramid and normal-force bounds, not measured forces or all physical constraints. Positive values indicate violation, negative values a margin. These bounds hold in $97.87\%$ of updates. The $99.5$th and $99.9$th percentiles are $0.0032$ N and $0.0116$ N, respectively, with a maximum of $2.44$ N.

The snapshots in Fig.~\ref{fig:real_world_exp_snapshots} further demonstrate locomotion on asphalt, flagstones, uneven soil, cobblestones, irregular stone paving, and dense grass. These experiments validate closed-loop DR-MPC beyond a flat laboratory surface.

\section{Conclusions}

This paper presented DR-MPC, an optimization-feasible formulation with a tailored IPM exploiting contact-aware input parameterization. For the evaluated implementations of the same DR-MPC formulation, it achieves median end-to-end speedups of $16.0\times$ over HPIPM and $4.4\times$ over OSQP with comparable simulation tracking. Outdoor experiments demonstrate nominal $100$ Hz onboard MPC over irregular terrain. The median time was $4.48$ ms, with $99.87\%$ of updates within budget and first-step optimized forces satisfying the evaluated bounds in $97.87\%$ of updates. Future work will extend dynamics relaxation to nonlinear MPC and other robotic platforms.










\bibliographystyle{IEEEtran}
\bibliography{ref}

\end{document}